%% file: UnElisa_arxiv.tex
\documentclass[11pt]{article}

\usepackage{latexsym,amsmath,amssymb,amsfonts,amsthm,bbm}
\usepackage[square]{natbib}
\usepackage[table]{xcolor} 
\usepackage[colorlinks=true,citecolor=blue]{hyperref}
\usepackage{enumerate}
\usepackage{graphicx}
\usepackage{graphics,enumerate}
\usepackage{fancybox}
\usepackage{algorithmicx}
\usepackage{algpseudocode}
\usepackage{subfigure}
\usepackage{mathrsfs}
\usepackage{bm}
\usepackage[lined,boxed,commentsnumbered]{algorithm2e}
\usepackage{tikz}
\usetikzlibrary{matrix}
\usepackage{multirow}
\usepackage{booktabs}
\newtheorem{theorem}{Theorem}
\newtheorem{proposition}{Proposition}

\newtheorem{lemma}{Lemma}

\newtheorem{rmk}{Remark}

\usepackage{chngcntr}
\usepackage{apptools}
\numberwithin{equation}{section}
\numberwithin{figure}{section}
\numberwithin{table}{section}
\numberwithin{theorem}{section}
\numberwithin{lemma}{section}
\numberwithin{corollary}{section}

\def \bE {\mathbb{E}}

\def \bR {\mathbb{R}}

\def \Acal {\mathcal{A}}
\def \Pcal {\mathcal{P}}

\def \Xcal {\mathcal{X}}
\def \Ycal {\mathcal{Y}}

\def \Ncal {\mathcal{N}}

\def \Gcal {\mathcal{G}}
\def \Vcal {\mathcal{V}}
\def \Ecal {\mathcal{E}}
\def \bbI {\mathbb{I}}

\def\mathbi#1{\textbf{\em #1}}

\def \beta {\bm{\eta}}

\newcommand{\bit}[1]{{\bf\textit{#1}}}

\newenvironment{eq}{\begin{equation}\begin{aligned}}{\end{aligned}\end{equation}}
\newenvironment{eq*}{\begin{equation*}\begin{aligned}}{\end{aligned}\end{equation*}}

\newenvironment{keywords}
{\bgroup\leftskip 20pt\rightskip 20pt \noindent{\bf Keywords:} }%
{\par\egroup\vskip 0.25ex}
\usepackage{footnotebackref}
\usepackage{setspace}

\definecolor{orange}{RGB}{250,80,000}

\begin{document}
\title{Unsupervised Ensemble Learning via Ising Model Approximation with Application to Phenotyping Prediction}

 \author{ Luwan Zhang$^1$, Tianrun Cai$^2$ \\
 \em $^1$Department of Biostatistics, Harvard T.H. Chan School of Public Health, Boston MA \\
  \em $^2$Division of Rheumatology, Immunology, and Allergy, Brigham and Women's Hospital, Boston, MA}
    
\date{}

\maketitle

\begin{abstract}
Unsupervised ensemble learning has long been an interesting yet challenging problem that comes to prominence in recent years with the increasing demand of crowdsourcing in various applications.  In this paper, we propose a novel method-- unsupervised ensemble learning via Ising model approximation (unElisa) that combines a pruning step with a predicting step. We focus on the binary case and use an Ising model to characterize interactions between the ensemble and the underlying true classifier. The presence of an edge between an observed classifier and the true classifier indicates a direct dependence whereas the absence indicates the corresponding one provides no additional information and shall be eliminated. This observation leads to the pruning step where the key is to recover the neighborhood of the true classifier. We show that it can be recovered successfully with exponentially decaying error in the high-dimensional setting by performing nodewise $\ell_1$-regularized logistic regression. The pruned ensemble allows us to get a consistent estimate of the Bayes classifier for predicting. We also propose an augmented version of majority voting by reversing all labels given by a subgroup of the pruned ensemble. We demonstrate the efficacy of our method through extensive numerical experiments and through the application to EHR-based phenotyping prediction on Rheumatoid Arthritis (RA) using data from Partners Healthcare System.
\end{abstract}

\vskip 10pt
	

\clearpage
\newpage

\baselineskip=24pt

\input{intro}

\section{Problem formulation}
We consider the following binary classification problem. Suppose we have $p$ binary classifiers of unknown reliability denoted as $\{f_s\}_{s=1}^p$ defined on a common domain $\Xcal$, i.e., $\forall s = 1, \ldots, p$
$$f_s: \Xcal \mapsto \{-1, +1\}$$ 
with each providing a prediction on a set of i.i.d. instances $D = \{x_i\}_{i=1}^n \subset \Xcal^n$. We use $f_s^{(i)}$ to denote the prediction given by $f_s$ on the i-th instance $x_i$. Let $f_0$ denote the underlying true classification rule on the domain $\Xcal$. Thus, $(f_0(X), f_1(X), \ldots, f_p(X))^{\top}$ forms a random vector on $\Ycal = \{-1, +1\}^{p+1}$, and the joint distribution of such a random vector on $\Ycal$ takes the form
\begin{eq}
\label{eq:jointpdf}
 P_{\theta^*}(f_0(x),..., f_p(x)) = \exp \Big\{\theta_{0}^*f_0(x) 
 + \sum_{0\leq s < t \leq p} \theta^*_{st}f_s(x)f_t(x) - A(\theta^*) \Big\}, \forall x \in \Xcal
\end{eq} 
where $\theta^*_{st} \in \bR$ encodes the strength of dependence between $f_s$ and $f_t$, and $\theta_0^* \in \bR$ determines the prior distribution of $f_0(X)$ since $P_{\theta^*}(f_0(x)= +1) = \frac{e^{2\theta_0^*}}{1+e^{2\theta_0^*}}, \forall x \in \Xcal$. In the following, we simply write $f_s$ for the random variable $f_s(X)$, where the randomness should be understood implicitly. Denote by $\widetilde{P}_{\theta^*}$ the marginal distribution of $(f_1,...,f_p)$. Let $\Gcal = (\Vcal, \Ecal)$ denote the associated graph, where $\Vcal=\{s\}_{s=0}^n$ and $\Ecal = \{ (s,t): |\theta^*_{st}| > 0, 0 \leq s < t \leq p \}$. $\forall s \in \Vcal$, define its neighborhood as well as its degree:
$$\Ncal_s:=\{t \in V: |\theta^*_{st}| > 0\}, \quad d_s := |\Ncal_s|$$ 
Let $f_{\Ncal_s}$ denote all classifiers corresponding to its neighborhood $\Ncal_s$. Obviously, $f_s \perp f_{\Vcal\backslash \Ncal_s}$ given $f_{\Ncal_s}$. 

Without loss of generality, we assume $f_1, \ldots, f_{d_0}$ are directly connected to $f_0$. These classifiers are viewed as experts and hence $\Ncal_0$ is the expert set.  We assume the graph $\Gcal$ has the following properties:
 \begin{itemize}
 	\item[($G1$)](\textbf{Identifiability of $\mathbi{f}_\mathbf{0}$)} There is a unique hidden node corresponding to $f_0$. It is most densely connected in the sense that consider the degree sequence $\{d_{(s)}\}_{s=0}^p$ sorted in a descending order, then $d_0 = d_{(0)}$ and $d_0 \geq d_{(1)} + 2$. 

 	\item[($G2$)](\textbf{Non-informativeness of an non-expert}) Any non-expert node is only allowed to access to the hidden node through at least one expert node in $\Ncal_0$. This indicates that such an non-expert classifier is simply redundant and can be removed from the ensemble.
 	
 	\item[($G3$)](\textbf{Separability of experts}) No edge is allowed between any pair among expert nodes, i.e. $\theta^*_{st} = 0, \forall (s,t) \subseteq \Ncal_0$. This indicates that after removing all non-expert classifiers, the remaining ones are conditionally independent given $f_0$.
 \end{itemize}
 
 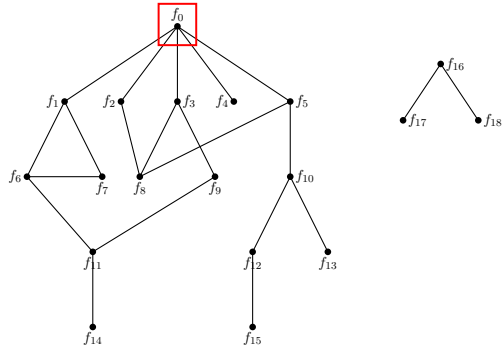
\begin{figure}[htbp]
 	\begin{center}
 		\resizebox{0.5\textwidth}{!}{\begin{minipage}{\textwidth}
 				\begin{center}
 					\begin{tikzpicture}
 					\draw[thick,-] (0,4) -- (-3,2);
 					\draw[thick,-] (0,4) -- (-1.5,2);
 					\draw[thick,-] (0,4) -- (0,2);
 					\draw[thick,-] (0,4) -- (1.5,2);
 					\draw[thick,-] (0,4) -- (3,2);
 					\draw[thick,-] (-3,2) -- (-4,0);
 					\draw[thick,-] (-3,2) -- (-2,0);
 					\draw[thick,-] (0,2) -- (-1,0);
 					\draw[thick,-] (-1.5,2) -- (-1,0);
 					\draw[thick,-] (3,2) -- (-1,0);
 					\draw[thick,-] (0,2) -- (1,0);
 					\draw[thick] (-4,0) -- (-2,0);
 					\draw[thick] (3,2) -- (3,0);
 					\draw[thick] (-4,0) -- (-2.25,-2);
 					\draw[thick] (1,0) -- (-2.25,-2);
 					\draw[thick] (3,0) -- (2,-2);
 					\draw[thick] (3,0) -- (4,-2);
 					\draw[thick] (2,-2) -- (2,-4);
 					\draw[thick] (-2.25,-2) -- (-2.25,-4);
 					\draw[thick] (7,3) -- (6,1.5);
 					\draw[thick] (7,3) -- (8,1.5);
 					\draw[thick, red, line width=1.5pt] (-0.5, 3.5) rectangle (0.5,4.6);
 					\fill (0,4) circle (2.5pt) node[above] {$f_0$};
 					\fill (-3,2) circle (2.5pt) node[left] {$f_1$};
 					\fill (-1.5,2) circle (2.5pt) node[left] {$f_2$};
 					\fill (0,2) circle (2.5pt) node[right] {$f_3$};
 					\fill (1.5,2) circle (2.5pt) node[left] {$f_4$};
 					\fill (3,2) circle (2.5pt) node[right] {$f_5$};
 					\fill (-4,0) circle (2.5pt) node[left] {$f_6$};
 					\fill (-2,0) circle (2.5pt) node[below] {$f_7$};
 					\fill (-1,0) circle (2.5pt) node[below] {$f_8$};
 					\fill (1,0) circle (2.5pt) node[below] {$f_9$};
 					\fill (3,0) circle (2.5pt) node[right] {$f_{10}$};
 					\fill (-2.25,-2) circle (2.5pt) node[below] {$f_{11}$};
 					\fill (2,-2) circle (2.5pt) node[below] {$f_{12}$};
 					\fill (4,-2) circle (2.5pt) node[below] {$f_{13}$};
 					\fill (-2.25,-4) circle (2.5pt) node[below] {$f_{14}$};
 					\fill (2,-4) circle (2.5pt) node[below] {$f_{15}$};
 					\fill (7,3) circle (2.5pt) node[right] {$f_{16}$};
 					\fill (6,1.5) circle (2.5pt) node[right] {$f_{17}$};
 					\fill (8,1.5) circle (2.5pt) node[right] {$f_{18}$};
 					
 					\end{tikzpicture} \end{center} \end{minipage}}
 		\caption{An illustrative example of a graph structure satisfying properties ($G1$)--($G3$). $f_0$ in the red square is unknown. The expert set $\Ncal_0 = \{1, 2, 3, 4, 5\}$. }
 		\label{fig: muti-layers}  
 	\end{center}
 \end{figure}
 
 Therefore, the underlying joint distribution given in (\ref{eq:jointpdf}) can be further decomposed into:
 \begin{eq}
 \label{eq:jointpdf2}
 P_{\theta^*}(f_0, \ldots, f_p) \propto \exp \left\{  \theta_0^* + \sum_{s=1}^{d_0} \theta^*_{0s}f_0f_s
 + \sum_{1\leq s < t \leq p} \theta^*_{st}f_sf_t \right\}
 \end{eq}
Since the prediction performance of $\{f_s\}_{s=1}^p$ on the domain $\Xcal$ is unknown, two natural questions are worth of asking:
\begin{itemize}
	\item[(1)] Among all availabel classifiers, who are experts? 
	
	\item[(2)] How to rely solely on experts in the ensemble to deliver a more accurate prediction rule? 
\end{itemize}
To answer (1), it is easy to see that it essentially amounts to estimate $\Ncal_0$, i.e., test $\theta_{0s}^* =0, \forall s \in \Vcal$. To address (2), it is in fact tightly related to (1) because the Bayes classifier denoted by $f_{B}(\cdot)$ , which minimizes the 0-1 misclassification error among all possible classifiers, is fully dependent on $f_{\Ncal_0}$. More explicitly, the Bayes classifier is
\begin{equation}
\label{eq:Bayes}
f_{B}(x) = sgn \left\{\sum_{s \in \Ncal_0} \theta^*_{0s}f_s(x) + \theta^*_0\right\}, \forall x \in \Xcal
\end{equation}
To solve the aforementioned two tasks, we propose the following two-step procedure: 
\begin{itemize}
	\item[(1)] Pruning step -- estimate the expert set denoted as $\widehat{\Ncal}_0$.
	\item[(2)] Predicting step -- estimate $\theta^*_{0s}, \forall s \in \widehat{\Ncal}_0$ denoted as $\widehat{\theta}^*_{0s}$, and plug in (\ref{eq:Bayes}) to estimate the Bayes classifier denoted as $\widehat{f}_B(\cdot)$.
\end{itemize}

\section{Pruning step}
In this section, we will elaborate in full details the pruning step to estimate $\Ncal_0$ using $\ell_1$-regularized Logistic regression based on the Ising model approximation to the marginal distribution $\widetilde{P}_{\theta^*}$.


\subsection{Ising model approximation to $\widetilde{P}_{\theta^*}$}
For any node $s \in \Vcal$, let $f_{\backslash s} := f_{\Vcal \backslash \{0, s\}}$. $\forall s \in \Vcal\backslash \{0, \Ncal_0\}$, 
\begin{equation}
\label{eq:cond-logit}
\frac{P_{\theta^*}(f_s=+1|f_{\backslash s})}{P_{\theta^*}(f_s =-1|f_{\backslash s)}} = \exp \left\{2\sum_{r \in \Ncal_s} \theta^*_{rs}f_r\right \}
\end{equation}
which indicates the conditional distribution of an non-expert $f_s$ given all other availabel classifiers follows a logistic model. With this observation, one can estimate $\Ncal_s$ by performing an $\ell_1$-regularized logistic regression of $f_s$ on $f_{\backslash s}$ following Ravikumar et al. in \cite{ravikumar2010high}. 
However, for any node $s \in \Ncal_0$, its conditional distribution $P_{\theta^*}(f_s|f_{\backslash s})$:
\begin{eq}
\label{eq:cond-not-logit}
\frac{P_{\theta^*}(f_s=+1|f_{\backslash s})}{P_{\theta^*}(f_s =-1|f_{\backslash s)}} =  \exp\left\{
2\sum_{r=d_0 +1}^p \theta_{rs}^* f_r \right\} \frac{e^{\theta^*_{0s} + L^*_s} + e^{-\theta^*_{0s} - L^*_s}}{e^{-\theta^*_{0s} + L^*_s} + e^{\theta^*_{0s} - L^*_s}}
\end{eq}
where $L^*_s := \theta_0^* + \sum_{\underset{t \neq s}{t \in \Ncal_0}} \theta^*_{0t}f_t$, no longer follows a logistic model. 

Although a logistic form is disabled in (\ref{eq:cond-not-logit}), Lemma \ref{lemma: main result} can show there always exists a unique Ising model that can best approximate $P_{\theta^*}(f_{\Ncal_0 \cup \Ncal_s \backslash 0})$ in terms of Kullback-Leibler divergence, and hence makes a logistic form re-eligible. In light of Lemma \ref{lemma: main result}, it is easy to show there is a unique Ising model that can best approximate $\widetilde{P}_{\theta^*}$ presented in Theorem \ref{thm: main thm}. 

\begin{lemma}
	\label{lemma: main result}
	For each node $s$ from $\Ncal_0$, there always exists a unique Ising model $Q^{s}_{\tilde{\theta}}(\cdot)$ that can best approximates the true distribution $P_{\theta^*}(f_{\Ncal_0 \cup \Ncal_s \backslash 0})$ in terms of Kullback-Leibler divergence.
	Furthermore, $Q^{s}_{\tilde{\theta}}(\cdot)$ takes the form
	$$ \exp\left \{ \sum_{\underset{r \neq t}{r,t \in \Ncal_0 \cup \Ncal_s \backslash 0}} \tilde{\theta}_{rt} f_rf_t - A(\tilde{\theta}) \right \}$$ in which
	\begin{eq}\label{eq: theta tilde}
		\tilde{\theta}_{rt} = \left \{\begin{array}{c c c} \theta^*_{rt} & \mbox{ if } & \{r,t\} \nsubseteq \Ncal_0 \\
		{1 \over 2} \log \left( \frac{e^{\theta^*_{0r} + \theta^*_{0t} + \theta_0^*}+ e^{-\theta^*_{0r} - \theta^*_{0t}-\theta_0^*} + e^{-\theta^*_{0r} - \theta^*_{0t} + \theta_0^*} + e^{\theta^*_{0r} + \theta^*_{0t} - \theta_0^*}}{e^{\theta^*_{0r} - \theta^*_{0t} + \theta_0^*}+ e^{-\theta^*_{0r} + \theta^*_{0t}-\theta_0^*} + e^{-\theta^*_{0r} + \theta^*_{0t} + \theta_0^*} + e^{\theta^*_{0r} - \theta^*_{0t} - \theta_0^*}}\right) & \mbox{if } &  (r, t) \subseteq \Ncal_0	\end{array} \right.
	\end{eq}

\end{lemma}

\begin{theorem}
	\label{thm: main thm}
	Suppose a random vector $(f_0, f_1, \ldots, f_p)^{\top} \in \Ycal$ follows the distribution given in (\ref{eq:jointpdf}), then there exists a unique Ising model $Q_{\tilde{\theta}}(\cdot)$ that best approximates $P_{\theta^*}(f_1, \ldots, f_p)$ marginalizing over $f_0$ in terms of Kullback-Leibler divergence. Furthermore,
	\begin{equation}
	\label{eq:Ising approx}
	Q_{\tilde{\theta}}(f_1,\ldots, f_p) = \exp \left\{ \sum_{1\leq s < t \leq p} \tilde{\theta}_{st}f_sf_t - A(\tilde{\theta}) \right\}
	\end{equation}
	in which $\tilde{\theta}_{st}$ takes the form given in (\ref{eq: theta tilde}).
\end{theorem}

Therefore, an immediate consequence from Theorem \ref{thm: main thm} is, $\forall s \in \Vcal \backslash \{0\}$, the best logistic model that can approximate its conditional distribution $P_{\theta^*}(f_s | f_{\backslash s})$ can be written in the form
\begin{equation}
\label{eq:approx-logit}
Q_{\tilde{\theta}}(f_s | f_{\backslash s}) = \frac{\exp\{2f_s\sum_{t \in \widetilde{\Ncal}_s} \tilde{\theta}_{st}f_t\}}{\exp\{2f_s\sum_{t \in \widetilde{\Ncal}_s} \tilde{\theta}_{st}f_t\} + 1}
\end{equation}

\begin{proposition}
	\label{prop:nonzero theta}
	Consider an Ising model approximation $Q_{\tilde{\theta}}(\cdot)$ given in Theorem \ref{thm: main thm}, $\forall \{s,t\} \subseteq \Ncal_0,  \; \tilde{\theta}_{st} \neq 0$. Furthermore, $ \; \tilde{\theta}_{st} > 0 \;$ if and only if $ \; \theta^*_{0s}\theta^*_{0t} > 0 \;$ and $ \; \tilde{\theta}_{st} < 0 \;$ otherwise.
\end{proposition}

\begin{figure}[htbp]
	\begin{center}
					\includegraphics[scale=0.3]{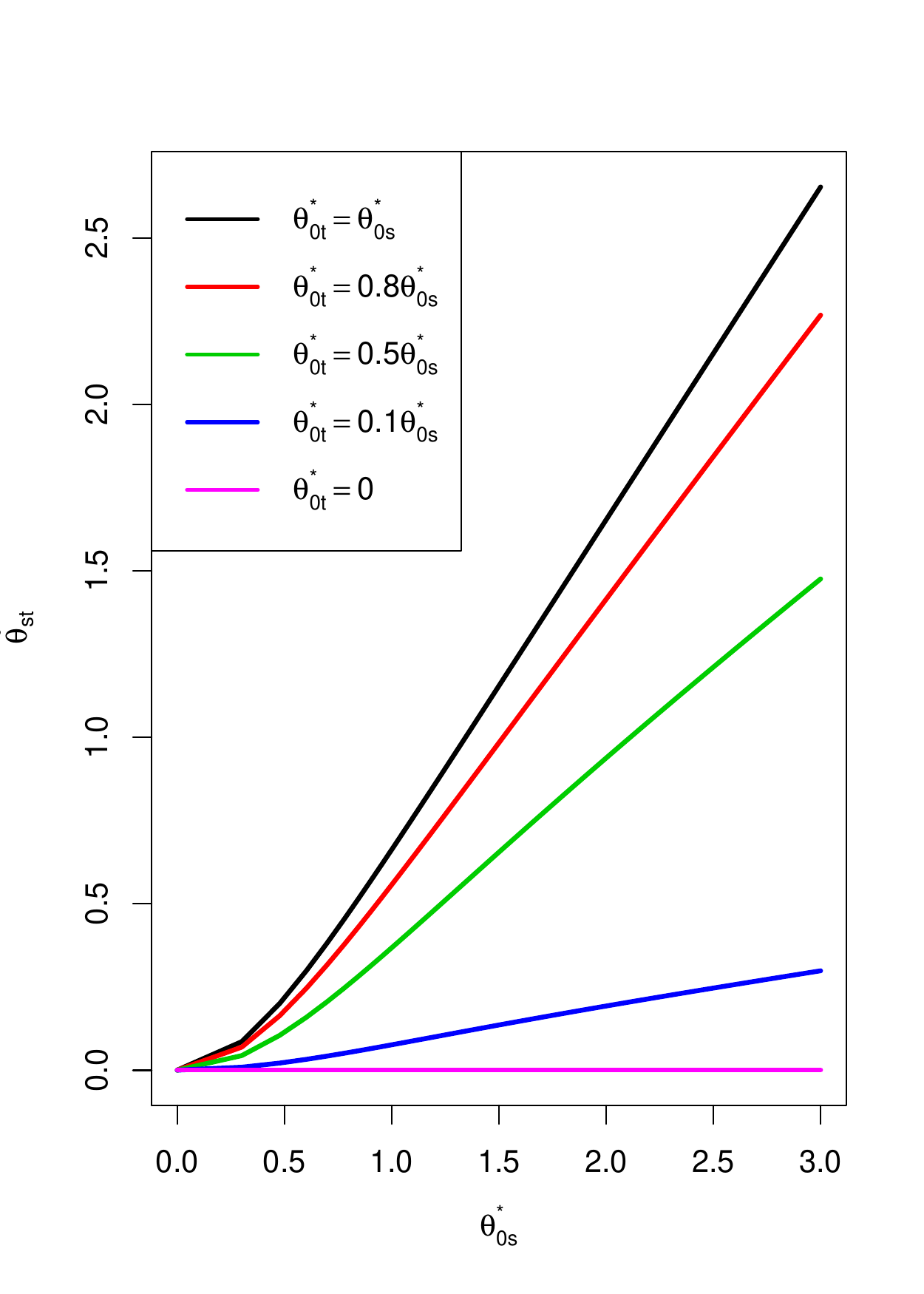}
                    \includegraphics[scale=0.3]{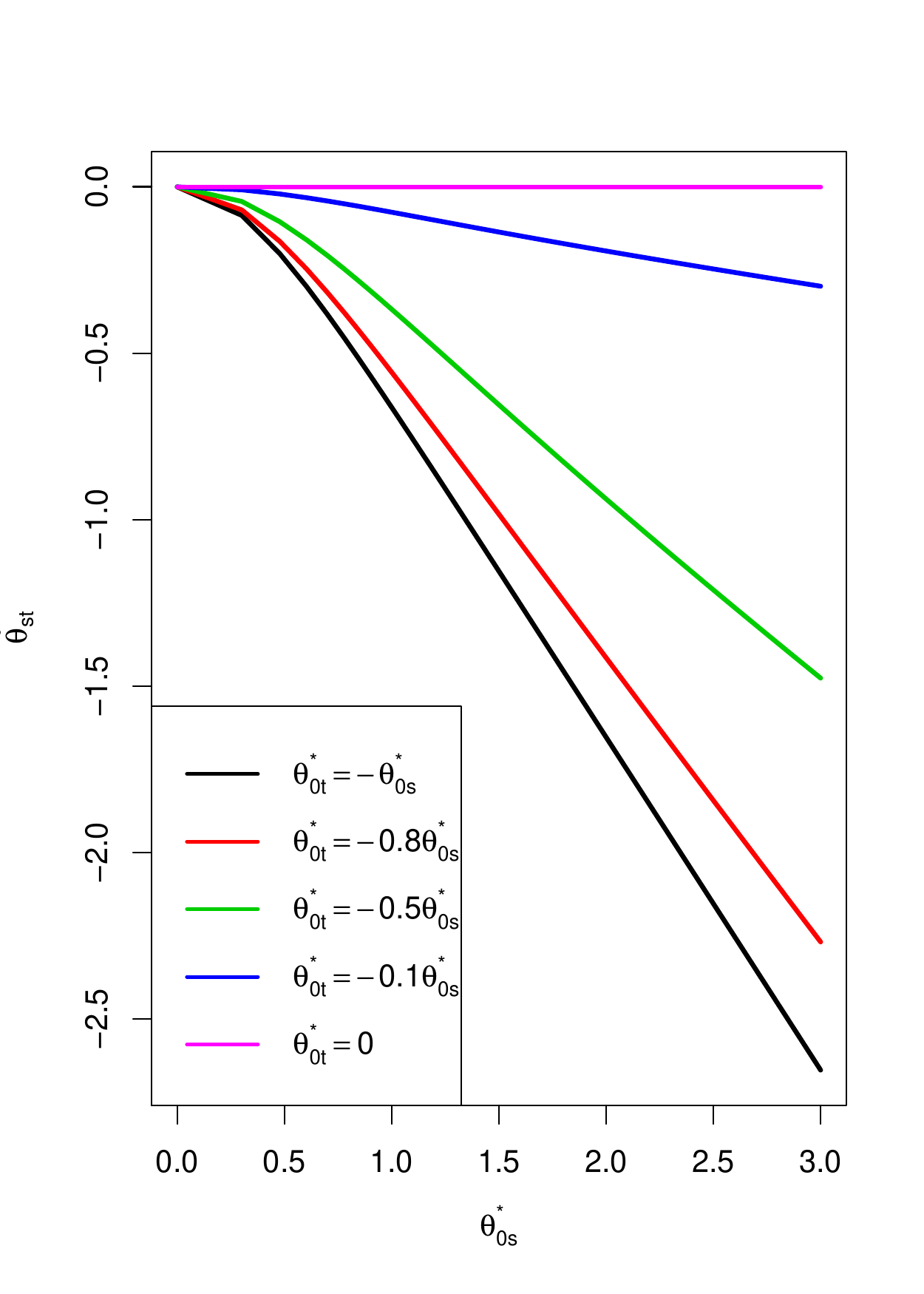}
		\caption{ Examples on the value of $\tilde{\theta}_{st}$ as a function of $ (\theta^*_{0s}, \theta^*_{0t}, \theta_0^*)$ given in Theorem \ref{thm: main thm}. The left panel corresponds to $\theta^*_{0s}\theta^*_{0t} > 0$ and the right corresponds to $\theta^*_{0s}\theta^*_{0t} < 0$. In both scenarios, set $\theta_{0}^*=0$.}
	    \label{fig: theta magnitude}
	\end{center}
\end{figure}

Proposition \ref{prop:nonzero theta} naturally induces the notion of neighborhood $\widetilde{\Ncal}_s$ with respect to $Q_{\tilde{\theta}}(\cdot)$ for each node $s \in \Vcal \backslash \{0\}$. Clearly, for an non-expert node, $\widetilde{\Ncal}_s = \Ncal_s$. In contrast, for an expert node, $ \widetilde{\Ncal}_s = \Ncal_s \cup \Ncal_0 \backslash \{0\}$, including its own neighborhood $\Ncal_s$ as well as siblings in $\Ncal_0$. For example, in Figure \ref{fig: muti-layers}, $\widetilde{\Ncal}_1 = \{2,3,4,5,6,7\}$.  Figure \ref{fig: theta magnitude} further demonstrates the joint effect of $(\theta^*_{0s}, \theta^*_{0t})$ on the value of $\tilde{\theta}_{st}$.
Intuitively, the more signals appearing on both edges simultaneously, the more capable $\tilde{\theta}_{st}$  of preserving them. The sign of $\tilde{\theta}_{st}$ can be used to infer whether there is an agreement on $f_s, f_t$.

\subsection{Neighborhood selection}
Now that there is a unique Ising model best approximating $\widetilde{P}_{\theta^*}$, for each node $s \in V\backslash \{0\}$, we can perform $\ell_1$-regularized logistic regression to select its neighborhood. The $\ell_1$-regularized regression is of the form
$$ \min_{\theta \in \bR^{p-1}} \left\{-\frac{1}{n} \sum_{i=1}^n \log Q_{\theta}(f^{(i)}_s | f^{(i)}_{\backslash s}) + \lambda_n \|\theta\|_1 \right\}$$ 
which amounts to
\begin{equation}
\begin{aligned}
\label{eq:ell-one}
\min_{\theta \in \bR^{p-1}} \left\{-\frac{1}{n} \sum_{i=1}^n \log \left\{\exp \left(\sum_{t \neq s}\theta_{st}f_t^{(i)} \right) + \exp \left(-\sum_{t \neq s}\theta_{st}f_t^{(i)} \right) \right\} - \sum_{t \neq s} \theta_{st}\widehat{\mu}_{st} + \lambda_n \|\theta\|_1 \right\} 
\end{aligned}
\end{equation}
where $\widehat{\mu}_{st} = {1 \over n}\sum_{i=1}^n f_s^{(i)}f_t^{(i)}$.
Note that the objective function (\ref{eq:ell-one}) is convex but not differentiable, due to the $\ell_1$-penalty term. However, Ravikumar et al in \cite{ravikumar2010high} showed that the minimizer $\widehat{\theta}$ is always achievable and unique under the regime of interest. Now we can use the edge potential estimate $\widehat{\theta}$ to reconstruct the corresponding neighborhood
$$\widehat{\Ncal}_s := \{t \in \{1,2,...,p\}\backslash s \; | \; |\widehat{\theta}_{st}| > 0\} $$

\subsection{Neighborhood selection correctness}
In this section, we mainly discuss the correctness of $\widehat{\Ncal}_s$ obtained by peforming $\ell_1$-regularized logistic regression. To establish some theoretical results, some additional assumptions are necessary, most of which are imposed on $Q_{\tilde{\theta}}(\cdot)$. More specifically,  $\forall s =1,\ldots, p$, consider its Fisher information matrix w.r.t $Q_{\tilde{\theta}}(\cdot)$
$$I_s := - \bE_{\tilde{\theta}} \left [\nabla^2 \log Q_{\tilde{\theta}}(f_s | f_{\backslash s})\right]
= \bE_{\tilde{\theta}} \left [h_{\tilde{\theta}}(f) f_{\backslash s}f_{\backslash s}^{\top}\right]$$
where $h_{\tilde{\theta}}(f) = \frac{4\exp(2f_s\sum_{t \neq s} \tilde{\theta}_{st}f_t)}{(\exp(2f_s\sum_{t \neq s} \tilde{\theta}_{st}f_t)+1)^2}$. For notation convenience, we temporarily drop the subscript denoting the node index and use $\Ncal$ to stand for the corresponding neighborhood and denote by $\Ncal^c$ its complement. The following assumptions hold for each $s = 1,\ldots, p$.
\begin{itemize}
	\item[($A1$)] There exists a pair of positive constants $(C_{min},\tilde{d}_{max})$ such that the Fisher information matrix $I$ restricted on its neighborhood $\Ncal$ denoted as $I_{\Ncal \Ncal}$ satisfies $$\lambda_{min}(I_{\Ncal \Ncal}) \geq C_{min}, \quad
	\lambda_{max}(\bE_{\tilde{\theta}}[f_{\backslash s}f_{\backslash s}^{\top}] ) \leq \tilde{d}_{max}$$
	\item[($A2$)] There exists an $\alpha \in (0, 1]$ such that $\|I_{\Ncal^c \Ncal}(I_{\Ncal \Ncal})^{-1}\|_{\infty} \leq 1-\alpha$.
\end{itemize}
We further denote the edge set w.r.t. $Q_{\tilde{\theta}}$ by $\tilde{\Ecal}$ and the degree for each node $s$ by  $\tilde{d}_s$. Clearly, $\tilde{d}_s = d_s + d_0 - 1, \forall s \in \Ncal_0$ and $\tilde{d}_s = d_s$ for the rest. Let $\tilde{d}_{max} := \max_{s \in \{1\ldots p\}} \tilde{d}_s$, and $\tilde{\theta}_{min} := \min_{(s,t) \in \tilde{\Ecal}} |\tilde{\theta}_{st}|$ 

\begin{theorem}
	\label{thm:neighborhood correctness}
	Consider an Ising model $Q_{\tilde{\theta}}(\cdot)$ given in (\ref{eq:Ising approx}) such that $(A1)$ and $(A2)$  are satisfied, let the regularization parameter $\lambda_n \geq \frac{16\alpha}{1-\alpha}\sqrt{\frac{\log p}{n}}$, $n > L\tilde{d}_{max}^3\log p$ for some positive constants $L$ and $K$, independent of $(n,p,\tilde{d}_{max})$, and $ \tilde{\theta}_{min}= \Omega\left(\sqrt{\frac{\tilde{d}_{max}\log p}{n}}\right)$, with probability at least $1-2\exp(-K\lambda_n^2n)$,
	$$\forall s \in \Ncal_0, \;  \widehat{\Ncal}_s = \widetilde{\Ncal}_s $$
	$$\forall s \in V\backslash \{0, \Ncal_0\}, \; \widehat{\Ncal}_s = \Ncal_s $$
	In addition, $\forall (s,t) \in \tilde{\Ecal}$, the sign of $\tilde{\theta}_{st}$ can be correctly recovered. 
\end{theorem}
\begin{rmk}
	Theorem \ref{thm:neighborhood correctness} is a direct result following the proof given in Ravikumar et al in \cite{ravikumar2010high} in which more details can be referred to.
\end{rmk}

\subsection{Reconstruction of $\Ncal_0$}
After recovering all neighborhoods $\widetilde{\Ncal}_s, \forall s = 1, \ldots, p$, a natural question is how to continue to reconstruct $\Ncal_0$. 

\begin{theorem}
	\label{thm: N_0 recover}
	Consider a graph $G$ satisfying the aforementioned properties ($G1$)--($G3$), and $\{\widetilde{\Ncal}_s\}_{s=1}^p$ denote a sequence of neighborhoods w.r.t $Q_{\tilde{\theta}}(\cdot)$. Furthermore, a node $s$ is said to be a \bit {knot} if $s$ is the only intersection of all its neighbor's neighborhoods, that is, $A_s=s$, where $A_s := \cap_{r \in \widetilde{\Ncal}_s} \widetilde{\Ncal}_r, s=1, \ldots, p$. Consider the collection of such knots denoted by $\Acal$, in which each knot $s$ is also associated with an index $i_s$ storing the position of $s$ in the ordered sequence $|\widetilde{\Ncal}_{(1)}| \geq |\widetilde{\Ncal}_{(2)}| \geq \ldots \geq |\widetilde{\Ncal}_{(|\Acal|)}|$, then
	$$\Ncal_0 = \{s \in \Acal: |\widetilde{\Ncal}_{s}| \geq i_s-1 \}$$
\end{theorem}

Theorem \ref{thm: N_0 recover} immediately suggests a procedure to recover $\Ncal_0$ which has been detailed in Algorithm \ref{alg:Ncal_0}.

 \begin{algorithm}[htbp]
 	\caption{Pruning step -- Reconstruction of $\Ncal_0$}
 	\label{alg:Ncal_0}
 	\KwData{$\{(f^{(i)}_1,\cdots, f^{(i)}_p)^{\top}\}_{i=1}^n$}
 	\KwResult{$\widehat{\Ncal}_0$}
 	
 	for $s = 1\ldots p$ do
 	\begin{itemize}
 		\item Perform $\ell_1$-regularized regression given in (\ref{eq:ell-one}).
 		\item Output its estimated neighborhood $\widehat{\Ncal}_s$
 	\end{itemize}
 	
 	Initialize $\Acal, \widehat{\Ncal}_0 = \emptyset$
 	
 	for $s = 1\ldots p $:
 	\begin{itemize}
 		\item $A_s := \cap_{r \in \widehat{\Ncal}_s} \widehat{\Ncal}_r$ 
 		\item If $A_s = s$, then 
 		$\Acal \leftarrow \Acal \cup s$
 	\end{itemize}
 	Sort the sequence $\{|\widehat{\Ncal}_s|\}_{s\in \Acal}$ in a descending order and associate each $s \in \Acal$ with an index $i_s$ defined in Theorem \ref{thm: N_0 recover}, then
    $$\widehat{\Ncal}_0 = \{s \in \Acal: |\widehat{\Ncal}_{s}| \geq i_s -1 \}$$	
 	 	
 	\KwRet{$\widehat{\Ncal}_0$} 
 \end{algorithm}
 
\section{Predicting step}
\subsection{Bayes classifier estimation}

\begin{figure}[htbp]
	\begin{center}
					\begin{tikzpicture}
					\draw[thick,-] (0,4) -- (-3,2);
					\draw[thick,-] (0,4) -- (0,2);
					\draw[thick,-] (0,4) -- (3,2);
					\draw[thick,dotted] (-2,2) -- (-1,2);
					\draw[thick,dotted] (1,2) -- (2,2);
					\draw[thick, red, line width=1.5pt] (-0.5, 3.5) rectangle (0.5,4.6);
					\fill (0,4) circle (2.5pt) node[above] {$f_0$};
					\fill (-3,2) circle (2.5pt) node[left] {$f_1$};
					\fill (0,2) circle (2.5pt) node[left] {$f_s$};					
					\fill (3,2) circle (2.5pt) node[right] {$f_{d_0}$};
					
					\end{tikzpicture} 
					\caption{The reduced graph structure }
					\label{fig: CIgraph}

\end{center}
\end{figure}
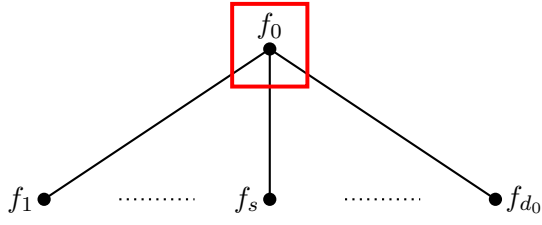

The recovery of $\Ncal_0$ has valuable spin-offs in the sense that only experts need to remain in the ensemble, reducing the graph to a tree structure with $f_0$ well separating all experts $f_{\Ncal_0}$ shown in Figure \ref{fig: CIgraph}.
Under a tree structure, it is easy to construct the Bayes classifier, which is known to be optimal among all possible classifiers to minimize the 0-1 misclassification error. The Bayes classifier in this case has the closed form 
$$f_{B}(\cdot) = sgn \left\{\sum_{s \in \Ncal_0} \theta^*_{0s}f_s(\cdot) + \theta_0^* \right\}$$ 
Our task now is to get a consistent estimate on $\theta^*_{0s}, s \in \Ncal_0$. To this end, let $\psi_s := P_{\theta^*}(f_sf_0 = + 1), s\in \Ncal_0$ and $\pi := P_{\theta^*}(f_0 = +1)$. According to the joint probability given in (\ref{eq:jointpdf}), some algebraic manipulations reveal the relationship between $(\theta_{0s}^*, \psi_s), s \in \Ncal_0$:

\begin{equation}
\label{eq:pair}
\begin{aligned}
\theta^*_{0s} &= {1 \over 2} \log \frac{\psi_s}{1-\psi_s}
\end{aligned}
\end{equation}
Moreover,
\begin{equation}
\label{eq:alpha0}
\theta^*_0 = {1\over 2}\log {\pi \over {1-\pi}}
\end{equation}

Denote the hidden random variable indicating the underlying label for each instance $x_i$ by $Z^{(i)}=\bbI(f^{(i)}_0 = +1), i=1, \ldots, n$, the complete data likelihood $L(\Theta)$ is
\begin{eq}
\prod_{i=1}^n & \left\{ Z^{(i)}\pi \prod_{s \in \Ncal_0} \psi_j^{\frac{f_s^{(i)}+ 1}{2}}(1-\psi_s)^{\frac{1 -f_s^{(i)}}{2}} + (1 - Z^{(i)})(1-\pi) \prod_{s \in \Ncal_0} (1-\psi_s)^{\frac{f_s^{(i)} + 1}{2}}\psi_s^{\frac{1 -f_s^{(i)}}{2}}\right \}
\end{eq}	

The maximization on the log-likelihood $\log L(\Theta)$, in general, is known to be very difficult due to its non-concavity. Therefore, the EM algorithm introduced in 
\cite{dempster1977maximum} by Dempster et al.(1977) is often alternatively used which essentially maximizes in the coordinate ascent fashion over the lower bound of $\log L(\Theta)$ defined as:
\begin{equation}
\label{eq:lowerbound}
\begin{aligned}
\sum_{i=1}^n \sum_{s\in \Ncal_0} \left \{Z^{(i)}\log \left [\psi_s^{\frac{f_s^{(i)} + 1}{2}}(1-\psi_s)^{\frac{1 -f_s^{(i)}}{2}}\right ] 
+ (1 - Z^{(i)}) \log \left [(1-\psi_s)^{\frac{f_s^{(i)} + 1}{2}}\psi_s^{\frac{1 -f_s^{(i)}}{2}}\right]  \right\} \\
 + \sum_{i=1}^n \left (Z^{(i)}\log \pi + (1-Z^{(i)})\log(1-\pi)\right)
\end{aligned}
\end{equation}

Define the posterior probability of being +1 as
\begin{equation}
\label{eq:post}
\tau_i := P(f_0^{(i)} = +1 | f^{(i)}_{\Ncal_0}) = \frac{\exp \left\{2\left(\theta^*_0 + \sum_{s\in \Ncal_0} \theta^*_{0s}f_s^{(i)}\right)\right\}}{1 + \exp \left\{2\left(\theta_0^* + \sum_{s \in \Ncal_0} \theta_{0s}^*f_s^{(i)}\right)\right\}}
\end{equation}

In E-step, each $Z^{(i)}$ is updated by its expectation $\tau_i$. In M-step, 
the MLE of $\psi_s$ and $\pi$ are given by
\begin{equation}
\label{eq:mle}
\begin{aligned}
\widehat{\psi}_s = {1 \over 2} + {1 \over n}\sum_{i=1}^n (\tau_i - {1 \over 2}) f_s^{(i)}, \quad \widehat{\pi} = \frac{\sum_{j=1}^n \tau_i}{n}
\end{aligned}
\end{equation}
Based on (\ref{eq:pair}) and (\ref{eq:alpha0}), we can get the estimation on $\theta^*_{0s}$ and $\theta_0^*$. The whole procedure is summarized in Algorithm \ref{alg:em}.

\begin{algorithm}
	\caption{Predicting step -- Bayes classifier estimation}
	\label{alg:em}
	\KwData{$(f^{(i)}_{\Ncal_0})^{\top}\}_{i=1}^n$}
	\KwResult{$\{\widehat{f}_B^{(i)}\}_{i=1}^n$}
	
	Initialization:  $\theta_0^*, \theta^*_{0s}, s \in \Ncal_0$ 
	\Repeat{\emph{(\ref{eq:lowerbound}) converges}} { 
		E Step -- 
		\begin{itemize}
			\item update $\tau_i$ by (\ref{eq:post})
		\end{itemize} 
		M Step -- 
		\begin{itemize}
			\item $\pi \leftarrow \frac{\sum_{j=1}^n \tau_i}{n}$
			\item $\widehat{\psi}_s \leftarrow {1 \over 2} + {1 \over n}\sum_{i=1}^n (\tau_i - {1 \over 2}) f_s^{(i)}$  
			\item update $\widehat{\theta}_{0s}$ by (\ref{eq:pair}) and $\widehat{\theta}_0$ by (\ref{eq:alpha0})
		\end{itemize}}
		for $i = 1:n$ 
		\begin{itemize}
			\item[] $\widehat{f}_B^{(i)} \leftarrow sgn\left(\widehat{\theta}_0 + \sum_{s \in \Ncal_0} \widehat{\theta}_{0s}f_s^{(i)}\right)$ 
		\end{itemize}
		\KwRet{$\{\widehat{f}_B^{(i)}\}_{i=1}^n$} 
	\end{algorithm}

\subsection{$\Ncal_0$ partitioning and augmented majority voting}

$\Ncal_0$ \textbf{partitioning}: an expert node identity thusfar has been simply determined by whether $\theta^*_{0s} $ is 0. Moving forward, the sign of $\theta^*_{0s}$ can indeed reveal whether the node is positive or negative, where positive/negative is measured by $\psi_s = P_{\theta^*}(f_sf_0 = + 1) =\frac{e^{2\theta^*_{0s}}}{1+e^{2\theta^*_{0s}}}$ being strictly greater/less than 0.5. In line with one's intuition, a positive expert spells its action exhibiting more compliance with $f_0$ while a negative expert goes to the opposite. Proposition \ref{prop:nonzero theta} has shed light on the partitioning principle---any two expert nodes with a positive $\tilde{\theta}_{st}$ would go to the same group and vice versa.  

\noindent \textbf{Augmented majority voting}:  in addition to distinguishing positive and negative experts, the partitioning on $\Ncal_0$ actually leads to a nonparametric alternative to the Bayes classifier for predicting. More specifically, with the assumption that there are more positive experts than their negative peers,  the negative group can be identified as the smaller-sized group and convert to the positive by reversing labelings. Then the majority voting policy can be applied on this twisted dataset.

\section{Numerical experiments}

\subsection{Simulations}
In this section, we describe some experimental results to evaluate the performance of our proposed two-step unsupervised ensemble learning method, which includes the pruning step(Algorithm \ref{alg:Ncal_0}) and the subsequent predicting step(Algorithm \ref{alg:em}). 

Specifically, we considered a series of Ising models in which $\theta_{0s}^* =  1/-1, \forall s \in \Ncal_0$ with probability $0.7/0.3$ such that positive experts form the major party in $\Ncal_0$. Set $\theta^*_0 = 0$ such that $P_{\theta^*}(f_0(x) = +1) = 0.5, \forall x \in \Xcal$. Moreover, defining $\Big|{\theta^*_{0s} \over \theta_{st}^*}\Big|, \forall s\in \Ncal_0, t \in \Ncal_s$ as the signal-to-noise ratio (SNR) in this context , we considered three levels of ANR: high, medium, low, in which $\theta^*_{st} = \pm 0.25 , \pm 0.5, \pm 1$ with equal probability respectively. To fully investigate the performance of our method, experiments were performed under various scaling scenarios of $(n,p,d_0)$. In particular, $p \in \{25, 49, 81\}$, and $d_0 \in \{\log p, \sqrt{p}, p/4\}$. We set $n = 30 \tilde{d}_{max}\log p$ and the regularization parameter $\lambda_n = \sqrt{\frac{\log p}{n}}$. Given the distribution of the form in (\ref{eq:jointpdf}), we generated random data by Gibbs sampling. To improve the data quality, we collected every $2(p+1)$th sample after the first 1000 iterations. For each specific combination of $(n, p, d_0)$, 200 independent trials were performed and results were averaged over these trials. We examined the performance of our proposed pruning step by two commonly used metrics-- Hit Rate and Precision, where Hit Rate is defined as the portion among elements in $\Ncal_0$ that have been successfully recovered, and Precision corresponds to the portion among all the selected that are truly in $\Ncal_0$. 
\begin{equation}
\label{eq: two metrics}
\mbox{Hit Rate} := \frac{|\widehat{\Ncal}_0 \cap \Ncal_0|}{|\Ncal_0|}, \; \; \mbox{Precision } := \frac{|\widehat{\Ncal}_0 \cap \Ncal_0|}{|\widehat{\Ncal}_0|}
\end{equation}

To examine the prediction performance, our method was compared to several other existing methods, including majority voting (MV), the classical Dawid-Skene estimator (DS), and the more recently developed spectral meta-learner (SML) \citep{parisi2014ranking}, as well as its extension to the dependence case via latent variables (SML-Latent) \citep{jaffe2016unsupervised}. 
All results were presented in Table \ref{tb:snr high}--Table \ref{tb:snr low}. 
Overall our predictive performance using both the Bayes classier and the augmented majority vote wins over the others, and the more noises, the larger margin. As $d_0$ grows from being logarithmic to linear in $p$, our method enjoys a consistently high performance, despite occasional mediocre performances with the high noise level and $d_0$ linear in $p$. In fact, the high performance in the pruning plays an irreplaceable role in escorting the predicting to the success.  Although the SML based methods yielded competing performances, their prescences were not stable and relatively sporadic. With SNR increasing, interactions between non-experts become significant to the degree that the conditional independence assumption between non-experts given their respective latent variables in SML-Latent method no longer holds, and as a consequence its performance becomes less satisfactory. 


\begin{table}[htbp]
		\resizebox{0.8\textwidth}{!}{\begin{minipage}{\textwidth}\begin{center}
			\begin{tabular}{c|c|c|c|c|c|c|c|c|c} 
			\hline \hline
			$p$ & $d_0$ & Hit Rate & Precision & unElisa-Bayes & unElisa-AMV & SML & SML-Latent& DS & MV  \\ \hline
            \multirow{3}{*}{25} & $\log p$ & 0.986  & 0.987 & 0.942 & 0.946 & 0.647 & 0.581 & 0.513 & 0.607  \\ 
            & $\sqrt{p}$ & 0.998 & 1 & 0.981 & 0.981 &0.409 & 0.221 & 0.459 & 0.429 \\ 
            & $ p/4 $ & 0.986 & 0.996 & 0.994 & 0.994 & 0.978 & 0.790 & 0.568 & 0.732 \\ 
            \hline
            \multirow{3}{*}{49}& $\log p$ & 0.984  & 0.991 & 0.960 & 0.958 & 0.728 & 0.731 & 0.509 & 0.599 \\  
            & $\sqrt{p}$ & 0.984 & 0.989 & 0.994 & 0.994 & 0.988 & 0.968 & 0.566 & 0.787   \\ 
            & $ p/4 $ & 0.958 & 0.995 & 0.994 & 0.997 &0.525 & 0.909 & 0.532 & 0.642 \\ 
            \hline
            \multirow{3}{*}{81}& $\log p$ & 0.984  & 0.992 & 0.984 & 0.984 & 0.692 & 0.674 & 0.509 & 0.617  \\ 
            & $\sqrt{p}$ & 0.991 & 0.998 & 0.998 & 0.998 & 0.282 & 0.927 & 0.531 & 0.609  \\ 
            & $ p/4 $ & 0.656 & 0.754 & 0.995 & 0.995 & 0.696 & 0.107 & 0.511 & 0.548  \\ 
            \hline
			\end{tabular}
			\caption{Performance summary table under high SNR. The first two columns correspond to the two metrics defined in (\ref{eq: two metrics}). The third to the last column is the prediction accuracy corresponding to our method using Bayes and AMV, SML, latent SML, Dawid-Skene, and majority vote respectively. }
			\label{tb:snr high}
			\end{center} \end{minipage}}
\end{table}

\begin{table}[htbp]
		\resizebox{0.8\textwidth}{!}{\begin{minipage}{\textwidth}
					\begin{tabular}{c|c|c|c|c|c|c|c|c|c} 
						\hline \hline
						$p$ & $d_0$ & Hit Rate & Precision & unElisa-Bayes & unElisa-AMV & SML & SML-Latent& DS & MV \\ \hline
						\multirow{3}{*}{25} & $\log p$ & 0.984  & 0.989 & 0.934 & 0.939 & 0.596 & 0.365 & 0.519 & 0.584  \\ 
						& $\sqrt{p}$ & 0.971 & 0.993 & 0.941 & 0.950&0.499 & 0.606 & 0.498 & 0.574 \\ 
						& $ p/4 $ & 0.901 & 0.959 & 0.989 & 0.986& 0.920 & 0.381 & 0.519 & 0.674  \\ 
						\hline
						\multirow{3}{*}{49}& $\log p$ & 0.982  & 0.995 & 0.959 &0.959& 0.728 & 0.577 & 0.500 & 0.612 \\  
						& $\sqrt{p}$ & 0.967 & 0.984 & 0.899 & 0.909&0.392 & 0.257 & 0.494 & 0.370   \\ 
						& $ p/4 $ & 0.804 & 0.891 & 0.996 & 0.990&0.952 & 0.891 & 0.572 & 0.720 \\ 
						\hline
						\multirow{3}{*}{81}& $\log p$ & 0.985  & 0.997 & 0.984 & 0.984 & 0.460 & 0.636 & 0.527 & 0.614  \\ 
						& $\sqrt{p}$ & 0.906 & 0.929 & 0.995 & 0.995& 0.064 & 0.824 & 0.492 & 0.468   \\ 
						& $ p/4 $ & 0.462 & 0.685 & 0.796 & 0.790& 0.852 & 0.781 & 0.509 & 0.507 \\ 
						\hline
					\end{tabular}
					\caption{Performance summary table under medium SNR. The first two columns correspond to the two metrics defined in (\ref{eq: two metrics}). The third to the last column is the prediction accuracy corresponding to our method using Bayes and AMV, SML, latent SML, Dawid-Skene, and majority vote respectively. }
					\label{tb:snr medium}
				 \end{minipage}}
		\end{table}
		
\begin{table}[htbp]
		\resizebox{0.8\textwidth}{!}{\begin{minipage}{\textwidth}
				
					\begin{tabular}{c|c|c|c|c|c|c|c|c|c} 
						\hline \hline
						$p$ & $d_0$ & Hit Rate & Precision & unElisa-Bayes & unElisa-AMV& SML & SML-Latent& DS & MV  \\ \hline
						\multirow{3}{*}{25} & $\log p$ & 0.978  & 0.986 & 0.946 &0.940& 0.772 & 0.548 & 0.533 & 0.665  \\ 
						& $\sqrt{p}$ & 0.921 & 0.985 & 0.936 & 0.938 & 0.497 & 0.350 & 0.458 & 0.418 \\ 
						& $ p/4 $ & 0.838 & 0.788 & 0.976 &0.979& 0.922 & 0.823 & 0.577 & 0.860 \\ 
						\hline
						\multirow{3}{*}{49}& $\log p$ & 0.942  & 0.980 & 0.955 &0.957& 0.529 & 0.413 & 0.506 & 0.647 \\  
						& $\sqrt{p}$ & 0.475 & 0.487 & 0.888 & 0.897&0.698 & 0.668 & 0.611 & 0.916   \\ 
						& $ p/4 $ & 0.481 & 0.685 & 0.773 & 0.794& 0.894 & 0.283 & 0.483 & 0.712 \\ 
						\hline
						\multirow{3}{*}{81}& $\log p$ & 0.963  & 0.988 & 0.981 & 0.981& 0.461 & 0.361 & 0.514 & 0.557 \\ 
						& $\sqrt{p}$ & 0.498 & 0.492 & 0.903 & 0.912&  0.901 & 0.218 & 0.527 & 0.691  \\ 
						& $ p/4 $ & 0.112 & 0.292 & 0.711 & 0.704 & 0 & 0.215 & 0.390 & 0.260  \\ 
						\hline
					\end{tabular}
					\caption{Performance summary table under low SNR. The first two columns correspond to the two metrics defined in (\ref{eq: two metrics}). The third to the last column is the prediction accuracy corresponding to our method using Bayes and AMV, SML, latent SML, Dawid-Skene, and majority vote respectively. }
					\label{tb:snr low}
				 \end{minipage}}
		\end{table}

\input{realdata}

\bibliography{refs.bib}

\input{appendix}

\end{document}

%% file: intro.tex
\section{Introduction}
In the area of machine learning, driven by the persistent quest for a better predictive performance on a given classification problem, numerous techniques have been proposed, among which ensemble learning is perhaps one most notable effort, that combines existing classifiers in hope to outperform any individual constituent. Many ensemble learning methods developed by far have enjoyed great popularity and success in various applications, such as Random Forest \citep{breiman2001random}, Boosting \citep{freund1996experiments,freund1997decision}, among others. A base classifier in the ensemble $f_s$ can be obtained from a wide variety of sources, either from human advisers or machine-based algorithms.  Despite the long-lasting remarkable performance, all these methods are in essence supervised living on the prior knowledge of true labels. However, for many reasons, data nowadays oftentimes come without true labels or not in a timely manner, for example, due to labor intensiveness, time constraints,  budget limits, confidential issues, privacy concerns. This surge of unlabled data hence becomes a main drive to develop high-quality unsupervised ensemble learning methods enabling accurate subsequent research delivery. One prominent effort is Electronic Health Record (EHR)-based phenotyping prediction, a data-driven approach inferring whether a patient has a certain disease using his or her EHR to overcome the current phenotypic data scarcity. Although it would be ideal to get gold-standard disease status labels, labor-intensive manual chart reviews, requiring up to months of human efforts, need thoroughly conducted by domain experts, which severely limits the ability to achieve high-throughput phenotyping and largely hampers the advancement in large-scale next-generation omics studies (NGOS) such as Phenome-Wide Association Studies (PheWAS) screening for associations between genomic markers and a diverse range of phenotypes. While EHR-based phenotyping prediction has proved be able to reproduce some previously established results using phenotypic data from traditional means \citep[e.g.]{ritchie2010robust, denny2010identification}, the yielded algorithms are not necessarily highly predictive across phenotypes in part due to a huge number of potentially irrelevant features collected in the EHR, which signifies the need of an ensemble method to reduce noise and enhance predicting power.

 Dawid and Skene in \citep{dawid1979maximum} pioneered a classical ensemble learning method built upon the assumption that base classifiers $\{f_s\}_{s=1}^p$ in the ensemble are conditionally independent given the true classier $f_0$. Connection of each $f_s$ to $f_0$ is characterized using a pair of probability parameters, representing the sensitivity and specificity respectively. With all parameters estimated in a closed form based on EM algorithm, the posterior probability makes the prediction. More recently, Parisi et al. in \citep{parisi2014ranking} proposed a spectral method by constructing a meta-learner that could be expressed in a linear form of all base learners. The method essentially leverages the rank-one structure on the covariance matrix of $\{f_s\}_{s=1}^p$ when the conditional independence is assumed. Jaffe et al. in \citep{jaffe2016unsupervised} later extended the spectral method into a dependence case where $\{f_s\}_{s=1}^p$  are allowed to be dependent through some unobserved latent variables. The covariance matrix as a consequence can be written as a convex combination of two rank-one matrices. Despite its more generalizability, the model requires the knowledge on the number of latent variables that is often unknown in practice. Additionally no direct interections can be inferred among classifiers .  Shaham et al. in \citep{shaham2016deep} considered a deep neural network to handle more complex dependence structures. To directly handle classifer dependencies, Donmez et al. in \citep{donmez2010unsupervised} utilized the mechanism of hierarchical log-linear models \citep{bishop1977book} and in particular considered a second-order log-linear model to capture pairwise interactions. However, the parameter space in their model expands quadratically as $p$ increases, resulting in a very computation-intensive task that they noted did not yield empirically much-improved prediction performance. 
 
To the best our knowledge, relatively few unsupervised ensemble learning methods have discussed the applicability to the high-dimensional setting where the ensemble size $p$ grows with the data size $n$. But this is no doubt a very important problem concerning both computational efficiency and statistical accuracy, in particular in step with the increasing prevalence of Crowdsourcing. In a generic Crowdsourcing setting, a series of tasks is completed by soliciting contributions from a large number of people termed as workers. Each worker acting as a base classifier provides his/her own answer on each task.  Answners would be collected from the crowd by the organizer to help decision making. Despite many substantial advantages, whether it be cost, time savings, or unlimited resources, Crowdsourcing inevitably suffers from a lot of noise since workers are not necessarily commissioned from a specific relevant group or professional organization. Most are indeed self-volunteered with no reliability guarantee. Recognizing experts among the crowd in this scenario is the key to prediction accuracy delivery.  Beyond that, utilizing the whole crowd would incur issues like running out of memory or substantial computation slow-down and even outage. Martinez-Muoz et al. in \citep{martinez2009analysis} investigated several pruning strategies to reduce the ensemble size using an idea of ordered aggregation in which the order is determined by the accuracy on the training set. Unfortunately, performance evaluation is hopeless without true labels, arising a barrier on ensemble pruning when it comes to the unsupervised setting.


In this paper, we propose a novel two-step unsupervised ensemble learning method via Ising model approximation (unElisa) to deal with a more complex dependence structure as well as to handle the high-dimenional setting. Inspired by \citep{donmez2010unsupervised}, we use an Ising model to characterize the joint distribution of the ensemble $\{f_s\}_{s=1}^p$ together with the true classifier $f_0$. A classical graphical model, Ising model has witnessed rich applications for its neat representation and high interpretability in a variety of domains, including statistical physics, natural language processing, image analysis, and spatial statistics, among others \citep{ising1925beitrag,manning1999foundations,woods1978markov,hassner1980use,cross1983markov,ripley2005spatial}. Ising model allows us to explicitly keep track of how each individual element in a community conforms its behavior to the neighbors to achieve synergy. The strength of an edge potential could reflect the dependence intensity between the node pair. To further distinguish different roles, we introduce three types of nodes--- a unique hidden node; expert nodes; and non-expert nodes. The hidden node corresponds to $f_0$ due to its unsupervised nature. An expert node is defined as a node directly connected to the hidden node whereas an non-expert node is defined as the opposite. The abscence of edge from an non-expert node to the hidden node suggests the corresponding classifier provides no additional information regarding $f_0$ and shall be eliminated to reduce the ensemble size. The pruning essentially enables a significant reduction on the parameter space, ensuring estimation consistency and preserving computation efficiency. The induced tree structure also allows for employment of the Bayes classifier that is known to be optimal in terms of minimizing the 0-1 misclassification error loss in the subsequent predicting step. 

We organize the paper as follows. In Section 2, we give a formal formulation on the problem of interest. We break down the learning task into two steps: pruning step to select experts, predicting step to make predictions using the reduced and denoised ensemble. In Section 3 we narrate in full details the pruning procedure by neighborhood selection based on Ising model approximation. The predicting step is elaborated in Section 4, in which two predicting paradigms are discussed, including the Bayes classification and augmented mojority voting.  In Section 5, we discuss some numerical results to show the performance of our proposed method.

%% file: realdata.tex
\subsection{Real data}
We further demonstrate the efficacy of our proposed method through the application to phenotyping prediction. An EHR-based phenotyping algorithm is a classification rule constructed using feature(s) extracted from a patient's EHR. For example one can use counts on an International Classification of Diseases, Ninth Revision (ICD-9) code to create ``silver-standard'' labels, which can be viewed as a bespoke ``probability'' of having the phenotype and then be converted to $\pm 1$ by thesholding, determined by domain knowledge or percentiles in the observed data. Similarly, features can be derived from a patient’s clinical narrative notes in the EHR database via natural language processing (NLP). One example of NLP features is the positive mentions of various related medical concepts appeared in each patient’s notes confirming the presence of the target phenotype. Patients with very high ICD-9 or NLP counts generally have the phenotype, while patients with extremely low counts are unlikely to have the phenotype. While ICD-9 and NLP counts are the two most reliable sources constructing a large pool of phenotyping algorithms shown in some recent studies \citep[e.g.]{liao2010electronic, ananthakrishnan2013improving, xia2013modeling, castro2015identification}, strategies for selecting a highly informative subset and determining appropriate thresholds remain largely untapped which yet plays a significant role in filtering out irrelevant ones that would otherwise disturb the overall performance. Additionally, these counts tend to be higher for patients with more healthcare utilization regardless of their true underlying phenotype status, weighing in as another factor affecting their individual predictive performance.  
  
In this paper, we focused on predicting Rheumatoid Arthritis (RA) phenotype using datamart from Partners Healthcare System. The RA datamart included the records of 46,568 patients who had at least one ICD-9 code of 714.x (Rheumatoid arthritis and other inflammatory polyarthropathies) or had been tested for anticyclic citrullinated peptide. The RA status was annotated by domain experts for a random sample of 435 patients, among whom 98 were diagnosed with RA (RA+) and the rest were not (RA-). To build the ensemble, we set the feature space within the sphere of ICD-9 codes ``714.x'' and NLP terms. To identify candidate medical concepts relevant to RA such as ''joint pain'', we processed articles from five knowledge sources including Medscape, Wikipedia, Merck Manuals Professional Edition, Mayo Clinic and MedlinePlus by named entity recognition software using Unified Medical Language System (UMLS). 1061 distinct concepts identified by more than half of the five knowledge sources were included to build a dictionary relevant to RA \citep{yu2016surrogate}. Note processing was performed using Narrative Information Linear Extraction (NILE) on patient clinical notes to search for positive concept mentions. For example, a concept ''joint pain” appeared in a sentence ``He denies joint pain” was considered a negative mention and omitted. 929 concepts were dropped which appeared in fewer than 5\% notes that mentioned RA. As a result, 142 features including 9 ICD-9 codes and 132 NLP terms were utilized.  We then  thresholded each feature at its percentile from 0.1 to 0.9 by 0.1 as well as performing a kmeans clustering using the full data. The area under the receiver operating characteristic curve (AUC) score on each generated classifier was calculated on the labeled subset. The final ensemble was made of all classifiers with AUC score between 0.6 and 0.8. 

\begin{figure}[htbp]
	\begin{center}
		\includegraphics[scale=0.8]{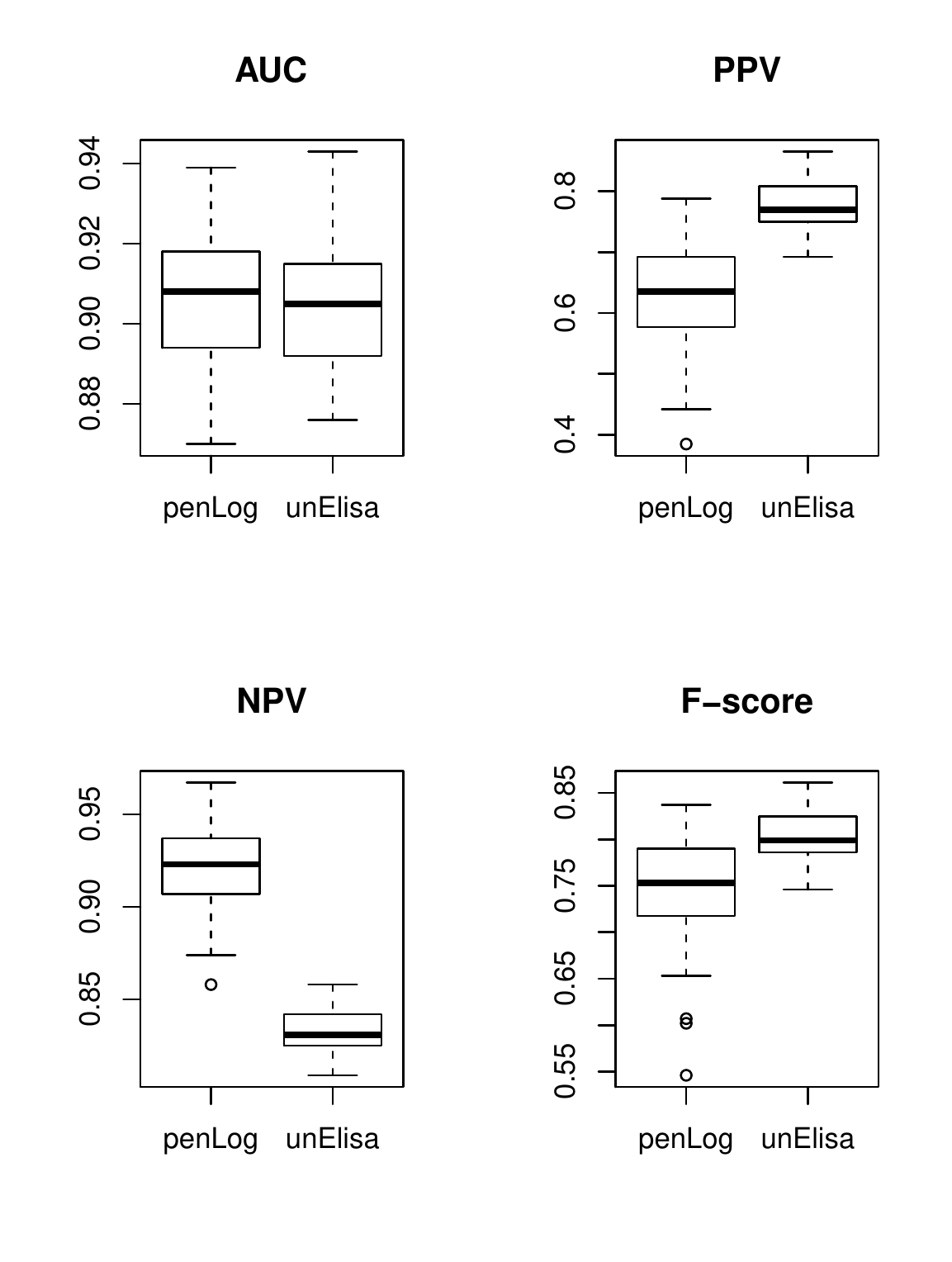}
		\caption{Boxplots of out-of-sample performance comparison on each metric.}
		\label{fig: RA}
	\end{center}
\end{figure}

Our algorithm can successfully identify highly predictive features among the crowd. For example, our algorithm selected in total 12 features including the main ICD-9 code ``714.0” and NLP concepts such as ``rheumatoid arthritis”, ``morning stiffness”, ``C-Reactive protein” and ``Immunosuppressive Agents” that are known highly clinically relevant to RA \citep{heidari2011rheumatoid}. On the other hand, concepts such as ``chest pain", ``diarrhea" and ``Influenza" which are not relevant to RA but automatically selected to build the dictionary for note processing were effectively discarded by our algorithm. 
For comparison, a penalized Logistics regression (penLog) was implemented using the same ensemble as covariates and gold-standard labels as the response. We then investigated each corresponding out-of-sample performance quantified by AUC, positive predictive value (PPV), negative predicted value (NPV) and the F-score \citep{fawcett2006introduction}, where
$$ \mbox{PPV} = \frac{\mbox{\# of true RA+}}{\mbox{\# of predicted RA+}}, \quad \mbox{NPV} = \frac{\mbox{\# of true RA-}}{\mbox{\# of predicted RA-}}, \quad \mbox{F-score} = \frac{2\mbox{PPV}*\mbox{NPV}}{\mbox{PPV}+\mbox{NPV}}$$
To this end, we ran 100 independent experiments, within each of which 200 random samples were used for training. To maintain the ratio of RA+/RA-, 46 out of 200 training samples were randomly selected from RA+ group. The training step aimed to select experts in $\Ncal_0$ which would be subsequently fed into the algorithm for testing on the rest 235 samples.  Table \ref{table: RA} and Figure \ref{fig: RA} summarize the out-of-sample performance in terms of AUC, PPV, NPV, and F-score, exhibiting that our method is able to behave nearly as well as its supervised counterpart.
\begin{table}[htbp]
	\begin{center}
		\begin{tabular}{c|c|c|c|c}
			\hline \hline 
			Method & AUC & PPV & NPV & F-score \\ \hline 
			penLog & 0.907(0.0149) & 0.635(0.0768) & 0.922(0.0218) & 0.749(0.0528)\\
			\hline 
			unElisa & 0.904(0.0147) & 0.774(0.0359) & 0.833(0.0104) & 0.802(0.0240) \\
			\hline \hline
		\end{tabular}
		\caption{Out of sample performance table. The table presents the mean value of each metric based on 100 independent experiments and the corresponding standard deviation in parenthesis.}
		\label{table: RA}
	\end{center}
\end{table}

%% file: appendix.tex
\appendix
\section{Appendix}

\subsection{Proof for Lemma \ref{lemma: main result}}
\begin{proof}
Let's consider a family of Ising models 
\begin{small}
\begin{eq}
\Pcal_{\theta} = & \left\{  \exp \left( \sum_{\underset{r \neq t}{r,t \in \Ncal_0 \cup \Ncal_s \backslash 0}} \theta_{rt}f_rf_t - A(\theta) \right) \right.  \\
& \left. \Bigg \vert \theta \in \bR^d, d =\frac{(d_s+d_0-1)(d_s+d_0-2)}{2}, A(\theta) < +\infty  \right\}  
\end{eq}	
\end{small}

The goal is to find a distribution $Q_{\tilde{\theta}} \in \Pcal_{\theta}$ that has the minimal Kullback-Leibler divergence $D_{KL}(P_{\theta^*}(f_{\Ncal_0 \cup \Ncal_s \backslash 0})||Q)$. 
\begin{small}
\begin{eq*}
&\min_{Q\in \Pcal_{\theta}} D_{KL}(P_{\theta^*}(f_{\Ncal_0 \cup \Ncal_s \backslash 0})||Q) \\&  \Longleftrightarrow  \min_{Q \in \Pcal_{\theta}} \sum_{y \in \{-1, +1\}^{d_s + d_0-1}} -p^*(y)\log q(y) \\
&\Longleftrightarrow  \min_{Q \in \Pcal_{\theta}} \sum_{y \in \{-1, +1\}^{d_s + d_0-1}} -p^*(y)\left(\sum_{\underset{r \neq t}{r, t \in \Ncal_0 \cup \Ncal_s \backslash 0}} \theta_{rt}f_rf_t - A(\theta) \right) \\
&\Longleftrightarrow  \min_{Q \in \Pcal_{\theta}} \left (A(\theta) - \sum_{\underset{r \neq t}{r, t \in \Ncal_0 \cup \Ncal_s \backslash 0}} \theta_{rt} \Bigg (p^*(f_rf_t=1)-p^*(f_rf_t=-1) \Bigg) \right)
\end{eq*} 
\end{small}
where we use $p^*(\cdot)$ as a shorthand for $P_{\theta^*}(f_{\Ncal_0 \cup \Ncal_s \backslash \{0\}})$.

Since $\Pcal_{\theta}$ is a subset of the regular exponential family, $A(\theta)$ is a convex function of $\theta$. Therefore the optimal solution $Q_{\tilde{\theta}}(\cdot)$ can be given by the stationary condition:
\begin{equation}
\label{eq:stationary}
\frac{\partial A(\theta)}{\theta_{rt}} = \bE_{p^*} [f_rf_t]
\end{equation}
where $\bE_{p^*} [f_rf_t] = p^*(f_rf_t=1)-p^*(f_rf_t=-1)$.

$\forall r \in \Ncal_s\backslash 0, \forall t \in \Ncal_0$,
$$\bE_{p^*} [f_rf_t] = \frac{e^{2\theta^*_{rt}}-1}{e^{2\theta^*_{rt}}+1}$$

$\forall (r, t) \subseteq \Ncal_0$,
$$\frac{p^*(f_rf_t=1)}{p^*(f_rf_t=-1)} = \frac{e^{a_1}+ e^{-a_1} + e^{a_2} + e^{-a_2}}{e^{a_3}+ e^{-a_3} + e^{a_4} + e^{-a_4}} \triangleq b_{st}$$
where $a_1 = \theta^*_{0s} + \theta^*_{0t} + \theta_0^*$, $a_2 = \theta^*_{0s} + \theta^*_{0t} - \theta_0^*$, $a_3 =\theta^*_{0s} - \theta^*_{0t} + \theta_0^*$, $a_4=\theta^*_{0s} - \theta^*_{0t} - \theta_0^*$.
$$\bE_{p^*} [f_rf_t] = \frac{b_{st} -1}{b_{st} + 1} \triangleq \mu_{st}$$ 
On the other hand, the moment matching conditions given by Wainwright and Jordan in \cite{wainwright2008graphical} reveals that
\begin{equation}
\label{eq:moment matching 2}
\frac{\partial A(\theta)}{\partial \theta_{rt}}\Bigg |_{\tilde{\theta}_{rt}} = \bE_{\tilde{\theta}} [f_rf_t] = \frac{e^{2\tilde{\theta}_{rt}}-1}{e^{2\tilde{\theta}_{rt}}+1}
\end{equation}
Therefore, combining (\ref{eq:stationary}) and (\ref{eq:moment matching 2}) together,
	$$\tilde{\theta}_{rt} = \theta^*_{rt}, \; \forall (r, t) \nsubseteq \Ncal_0$$
	$$\tilde{\theta}_{rt} = {1 \over 2} \log(b_{st}), \;  \forall (r,t) \subseteq \Ncal_0$$
	The uniqueness of $Q_{\tilde{\theta}}(\cdot)$ is naturally followed.
\end{proof}

\begin{proof}
	The proof basically goes through the same flow given in the proof of Lemma \ref{lemma: main result} and hence we dismiss the repetitive details by only providing a proving sketch. First, we start to build an Ising model family indexed by edge potential parameters $\theta \in \bR^d$, where $d = \frac{p(p-1)}{2}$. Next, by using the stationary condition (\ref{eq:stationary}) combined with the moment matching equations (\ref{eq:moment matching 2}), the desired result can be obtained.
\end{proof}

\subsection{Proof for Proposition \ref{prop:nonzero theta}}
\begin{proof}
	We first show $\tilde{\theta}_{st} \neq 0$. Consider a bivariate function $h(x,y) = e^{x+y} + e^{-x-y}+e^{x-y}+e^{y-x}$, for a fixed $y \in \bR, h(\cdot,y)$ is a symmetric and strictly convex function since $\frac{\partial^2 h}{\partial x^2} = (e^x+e^{-x})(e^y+e^{-y})>0$. To show $\tilde{\theta}_{st} \neq 0$, it is essential to show $h(\theta^*_{0s}+\theta^*_{0t}, \theta^*_0) \neq f(\theta^*_{0s}-\theta^*_{0t}, \theta^*_0)$. As $\theta^*_{0s}, \theta^*_{0t} \neq 0$, $\theta^*_{0s}+\theta^*_{0t} \neq \pm (\theta^*_{0s}-\theta^*_{0t})$, which ensures the argument inside the logarithm is not 1. Next, since $\tilde{\theta}_{st} > 0 \; \Longleftrightarrow  \; h(\theta^*_{0s}+\theta^*_{0t}, \theta^*_0) > h(\theta^*_{0s}-\theta^*_{0t}, \theta^*_0)$, $ \; \theta^*_{0s}\theta^*_{0t} > 0 \; \Longleftrightarrow \; |\theta^*_{0s}+\theta^*_{0t}| > |\theta^*_{0s}-\theta^*_{0t}|$, and the strict convexity of $h(\cdot, \theta_0^*)$ ensures $h(\theta^*_{0s}+\theta^*_{0t}, \theta^*_0) > h(\theta^*_{0s}-\theta^*_{0t}, \theta^*_0) \; \Longleftrightarrow \; |\theta^*_{0s}+\theta^*_{0t}| > |\theta^*_{0s}-\theta^*_{0t}|$, the proof is complete.
\end{proof}

\subsection{Proof for Theorem \ref{thm: N_0 recover}}
\begin{proof}
	We first show $\forall s \in \Ncal_0, s \in \Acal$. Obviously, $s \in A_s$. $\forall t \in \Ncal_0 \backslash \{s\}$, Proposition \ref{prop:nonzero theta} ensures $\tilde{\theta}_{st} \neq 0$ such that $s \in \widetilde{\Ncal}_{t}$. Therefore, $|\widetilde{\Ncal}_s| \geq d_0 -1$. Suppose there exist $r \neq s$ such that $r \in A_s$, then $r \notin \Ncal_0$ and $d_r \geq d_0 -1$ since it must appear in each neighbourhood $\widetilde{\Ncal}_t, \forall t \in \Ncal_0 \backslash s$, which contradicts with ($G1$). Therefore, $s \in \Acal$. Next, we need to show  $\forall s \in \Ncal_0, |\widetilde{\Ncal}_{s}| \geq i_s-1$. This is obvious due to the fact $|\widetilde{\Ncal}_{s}| \geq d_0-1$ and they correspond to the first $d_0$ largest $|\widetilde{\Ncal}_s|$. On the other side, if there exists a non-expert node $s \in \Acal$, as $|\tilde{\Ncal_s}|< d_0 -1$, then its corresponding index $i_s \geq d_0 + 1$, indicating $ |\widetilde{\Ncal}_{s}| < i_s-1$, therefore $\{s \in \Acal: |\widetilde{\Ncal}_{s}| \geq i_s-1 \} \subseteq \Ncal_0$.
\end{proof}